\documentclass[10pt,twocolumn,letterpaper]{article}

\RequirePackage{xspace}
\makeatletter
\DeclareRobustCommand\onedot{\futurelet\@let@token\@onedot}
\def\@onedot{\ifx\@let@token.\else.\null\fi\xspace}

\def\eg{\emph{e.g}\onedot}

\makeatother

\usepackage{cvpr}              
\usepackage{url}            
\usepackage{booktabs}       
\usepackage{amsfonts}       
\usepackage{nicefrac}       
\usepackage{microtype}      
\usepackage{xcolor}         
\usepackage{amsmath}
\usepackage{graphicx}
\usepackage{graphicx}
\usepackage{subcaption}
\usepackage{float}
\usepackage{xspace}
\usepackage{graphicx}

\usepackage{graphicx} 
\usepackage{wrapfig} 
\usepackage{graphicx}
\usepackage{colortbl}
\usepackage{xcolor}
\usepackage{booktabs}
\usepackage{subcaption} 
\usepackage{adjustbox} 
\usepackage{array}
\usepackage{overpic} 

\usepackage{pifont}       
\usepackage{bbding}       
\usepackage{fontawesome}
\usepackage{xspace}

\usepackage{amsthm,amsmath,amssymb}
 
\usepackage{mathrsfs}

\usepackage{makecell}
\usepackage{multirow}

\definecolor{adptorange}{RGB}{248, 205, 172}
\definecolor{cmpblue}{RGB}{189, 215, 238}
\definecolor{cmpblue}{RGB}{189, 215, 238}

\definecolor{our_red}{RGB}{232,157,160}
\definecolor{our_blue}{RGB}{136,206,230}
\definecolor{our_orange}{RGB}{246,200,168}
\definecolor{our_green}{RGB}{178,211,164}

\definecolor{important_token}{RGB}{205,146,16}

\renewcommand{\paragraph}[1]{\vspace{1mm}\noindent\textbf{#1}}
\newlength\savewidth
\usepackage{colortbl}
\usepackage{xcolor}
\usepackage{wrapfig}

\definecolor{bestcolor}{HTML}{CBDAEA}
\newcommand{\bestcell}[1]{\cellcolor{bestcolor}{#1}}

\newcommand{\tablestyle}[2]{\setlength{\tabcolsep}{#1}\renewcommand{\arraystretch}{#2}\centering\footnotesize}
\newcolumntype{x}[1]{>{\centering\arraybackslash}p{#1pt}}
\newcolumntype{y}[1]{>{\raggedright\arraybackslash}p{#1pt}}
\newcommand\hshline{\noalign{\global\savewidth\arrayrulewidth
  \global\arrayrulewidth 0.5pt}\hline\noalign{\global\arrayrulewidth\savewidth}}

\definecolor{cvprblue}{rgb}{0.21,0.49,0.74}
\usepackage[pagebackref,breaklinks,colorlinks,allcolors=cvprblue]{hyperref}

\title{\textit{A Stitch in Time Saves Nine}:\\
Small VLM is a Precise Guidance for Accelerating Large VLMs}

\author{
Wangbo Zhao$^{1}$\footnotemark[1]\quad
Yizeng Han$^{2}$\footnotemark[1]\quad
Jiasheng Tang$^{2,3}$\quad
Zhikai Li$^{1}$\quad
Yibing Song$^{2,3}$\\
Kai Wang$^{1}$\footnotemark[2]\quad
Zhangyang Wang$^{4}$\quad
Yang You$^{1}$\footnotemark[2]
\\
 $^{1}$National University of Singapore \quad
 $^{2}$DAMO Academy, Alibaba Group\\
 $^{3}$Hupan Lab \quad
 $^{4}$The University of Texas at Austin\\}

\begin{document}

\maketitle

\renewcommand{\thefootnote}{\fnsymbol{footnote}}
\footnotetext[1]{Equal contribution. Work done during an internship at DAMO Academy, Alibaba Group. wangbo.zhao96@gmail.com}
\footnotetext[2]{Corresponding author.}

\begin{abstract}
Vision-language models (VLMs) have shown remarkable success across various multi-modal tasks, yet large VLMs encounter significant efficiency challenges due to processing numerous visual tokens. A promising approach to accelerating large VLM inference is using partial information, such as attention maps from specific layers, to assess token importance and prune less essential tokens. However, our study reveals three key insights: (i) Partial attention information is insufficient for accurately identifying critical visual tokens, resulting in suboptimal performance, especially at low token retention ratios; (ii) Global attention information, such as the attention map aggregated across all layers, more effectively preserves essential tokens and maintains comparable performance under aggressive pruning. However, the attention maps from all layers requires a full inference pass, which increases computational load and is therefore impractical in existing methods; and (iii) The global attention map aggregated from a small VLM closely resembles that of a large VLM, suggesting an efficient alternative. Based on these findings, we introduce a \textbf{training-free} method, \underline{\textbf{S}}mall VLM \underline{\textbf{G}}uidance for accelerating \underline{\textbf{L}}arge VLMs (\textbf{SGL}). Specifically, we employ the attention map aggregated from a small VLM to guide visual token pruning in a large VLM. Additionally, an early exiting mechanism is developed to fully use the small VLM's predictions, dynamically invoking the larger VLM only when necessary, yielding a superior trade-off between accuracy and computation. Extensive evaluations across 11 benchmarks demonstrate the effectiveness and generalizability of SGL, achieving up to 91\% pruning ratio for visual tokens while retaining competitive performance. The code will be  publicly available at \url{https://github.com/NUS-HPC-AI-Lab/SGL}.
\end{abstract}
    
\section{Introduction}

Building on the notable success of  language models (LMs) \cite{brown2020language, touvron2023llama, jiang2023mistral, yang2024qwen2}, vision-language models (VLMs) have become a focal point of research. Most current VLMs \cite{zhu2023minigpt, liu2023improvedllava, liu2023llava, chen2024far, wang2024qwen2} integrate visual tokens from a vision encoder alongside textual tokens within an LM. However, such integration introduces significant inference overhead due to the sheer volume of visual tokens.

Visual token compression presents a compelling solution to improving the inference efficiency of VLMs. Recent works \cite{li2025llama, ye2024voco, ma2024visual} aim to condense the information in visual tokens into fewer tokens or parameters, but generally require training, introducing additional overhead. Training-free alternatives, such as token merging in visual encoders~\cite{bolya2022token, shang2024llava}, offer a lighter solution but may overlook essential vision-language interactions. To bridge this gap, \cite{chen2024image, zhang2024sparsevlm} employs \textit{partial information} from the LM \eg attention maps from specific layers, to prune less important visual tokens.  These approaches can ideally integrate seamlessly with existing VLMs without fine-tuning, showing promising effectiveness.

\begin{figure*}[!t]
    \centering
    \includegraphics[width=1\linewidth]{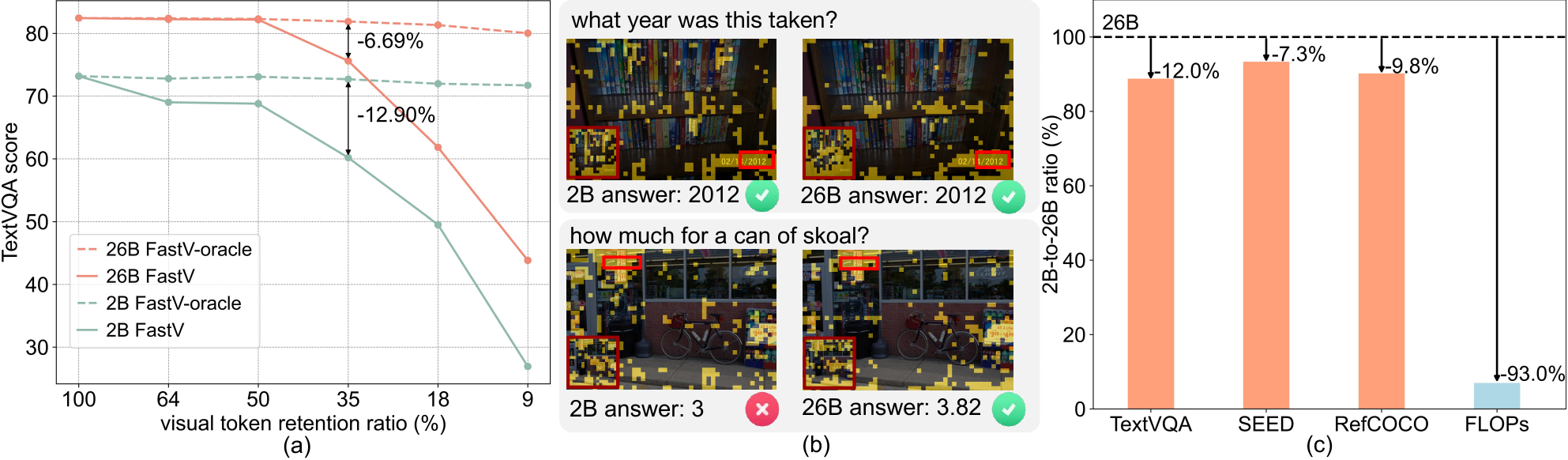}
    \vskip -0.1in
    \caption{{The motivation of our SGL.}
    \textbf{(a) A single-layer attention map is suboptimal compared to the global attention maps aggregated from all layers.} We take InternVL2~\cite{chen2024far} of 2B and 26B as representative examples. FastV~\cite{chen2024image} prunes visual tokens using the attention map from a single-layer, whereas FastV-oracle employs the aggregated attention map across all layers during inference. This approach allows for precise pruning of less significant visual tokens, maintaining performance with only 9\% of the tokens retained. 
    \textbf{(b) The small VLM exhibits a token retention pattern similar to the 26B model, preserving essential viusal tokens relevant to the answer, regardless of the answer correctness.}
    We drop 80\% less significant visual tokens and adopt \raisebox{0.25em}{\colorbox{important_token!50}{}} to mark those tokens with high attention scores. 
    Thumbnails employed in InternVL2~\cite{chen2024far} are presented in the left corner. 
    \textbf{(c) The performance gap between small and large VLM is minimal compared to their computation disparity.} The 2B model achieves competitive performance with significantly fewer FLOPs compared to the 26B one. This also validates our soundness of using a small model to guide early exiting and token pruning in the large one.} 
    \label{figure1}
       \vspace{-3mm}

\end{figure*}

\textit{But how effective is visual token pruning across varying retention levels?} To investigate this, we conduct an empirical study using FastV~\cite{chen2024image}, a representative method that assesses visual token importance based on a single-layer attention map from the LM. For comparison, we also include an oracle method aggregating attention maps from all LM layers. As shown in Figure~\ref{figure1} (a), FastV struggles to maintain accuracy when the token retention ratio falls below 35\%, whereas the oracle method remains competitive even with only 9\% of visual tokens retained. This highlights that \textit{the global information from the aggregated attention map accurately identifies essential visual tokens for VLM prediction}. 

However, retrieving attention maps from all layers requires a full inference pass. One cannot obtain a precise assessment of token importance before inference. This accounts for FastV using a single layer's attention as a proxy. To accurately distinguish essential vision tokens with minimal computation, we innovatively resort to a small model’s global attention maps aggregated from all layers. As shown in Figure~\ref{figure1} (b), the token retention pattern derived from the small VLM closely mirrors that of the large VLM and consistently preserves tokens relevant to the answer, \emph{regardless of output correctness}. This suggests that the small VLM’s overall attention map serves as a more precise proxy to guide visual token pruning in large VLMs.

Based on the above findings, we propose a \textbf{training-free} method, \underline{\textbf{S}}mall VLM \underline{\textbf{G}}uidance for \underline{\textbf{L}}arge VLMs (\textbf{SGL}), consisting of two insightful techniques. First, along the token pruning line, we develop \emph{Small VLM-Guided visual token Pruning (SGP)}. Given an input image and its text prompt (question), inference is first performed by a small VLM, in which a global attention map is aggregated from all layers. This global attention map enables the calculation of vision tokens' importance scores based on their interactions with prompt and generated tokens. The scores are then used to rank the visual tokens. Subsequently, the ranking results provide a guidance for pruning less important visual tokens in the large VLM, significantly reducing computation while preserving essential information for a correct answer.

It can be easily observed that the additional small VLM introduces some computational overhead. Fortunately, we notice that the performance gap between small and large VLMs is relatively minimal compared to their computation disparity (Figure~\ref{figure1} (c)). In other words, most ``easy'' questions could be correctly answered by the small VLM. This observation prompts us to make full use of the computation spent by the small VLM. To this end, we introduce \emph{Small VLM Early Exiting (SEE)}. Specifically, after obtaining the small VLM's prediction, we evaluate its decision score and directly terminate the inference pipeline without activating the large VLM if the score exceeds the threshold. With the adoption of a small VLM, SEE serves an effective approach complementary to SGP: \textcolor{blue}{(\emph{i})} The computation of the large VLM could be completely skipped for many ``easy'' questions (SEE); \textcolor{blue}{(\emph{ii})} When the large VLM is activated by demand, most unimportant visual tokens can be pruned based on the small VLM's aggregated attention map (SGP).

We demonstrate the effectiveness of SGL across 11 benchmarks, achieving up to 91\% pruning of visual tokens in VLMs such as InternVL2~\cite{chen2024far} in model sizes from 26B to 76B, while maintaining competitive performance. Moreover, our method can be seamlessly integrated with other VLMs \eg Qwen-VL \cite{wang2024qwen2} and LLaVa-OV~\cite{li2024llava} , highlighting its versatility in enhancing VLM efficiency across model architectures.

\section{Related Works}
\vspace{-1mm}
\paragraph{VLMs.}
Advancements in language models (LMs)~\cite{brown2020language, touvron2023llama, jiang2023mistral, yang2024qwen2} have driven significant progress in vision-language models (VLMs)~\cite{zhu2023minigpt, liu2023improvedllava, liu2023llava, chen2024far, wang2024qwen2}. Most VLMs use a visual encoder, such as ViT~\cite{dosovitskiy2020image}, to extract visual tokens connected to an LM via a projection layer, significantly increasing token length and leading to high computational and memory demands. For instance, LLaVa~\cite{liu2023llava} processes 576 tokens for a 336$\times$336 image resolution, while an enhanced version~\cite{liu2023improvedllava} processes 2304 tokens at higher resolutions. InternVL2~\cite{chen2024far} introduces up to 10,496 tokens through dynamic high-resolution techniques, and video understanding models~\cite{lin2023video, xu2024slowfast, zhang2024llavanextvideo} handle thousands of tokens across frames. 

Our method addresses the token overhead by using a small VLM to guide token reduction in larger VLMs, seamlessly applying to such models.

\vspace{-1mm}
\paragraph{Visual token compression.}
Compressing visual tokens is a promising approach to reduce computational and memory costs in transformer-based vision models. Methods such as token pruning~\cite{rao2021dynamicvit, liang2022not}, token merging~\cite{bolya2022token, chen2023diffrate}, and token skipping~\cite{zhao2024dynamic, meng2022adavit,han2021dynamic,han2024latency,zhao2024dydit} have been extensively studied for tasks like image classification and segmentation. In VLMs, methods like Q-Former~\cite{li2025llama}, token distillation~\cite{ye2024voco}, and parameter alignment~\cite{ma2024visual} compress visual tokens but often require additional training. Training-free techniques~\cite{bolya2022token, shang2024llava, chen2024image, zhang2024sparsevlm} merge or prune tokens based on LM attention maps, but approaches that merge tokens in the visual encoder~\cite{bolya2022token, shang2024llava} may miss key vision-language interactions. Other methods~\cite{chen2024image, zhang2024sparsevlm} prune tokens using attention maps from specific LM layers, which may fail to accurately capture essential tokens at low retention levels. 

In comparison, this work proposes leveraging the aggregated attention map across all layers of a small VLM to comprehensively rank token importance, guiding more effective pruning in a larger VLM.

\vspace{-1mm}
\paragraph{Model uncertainty/confidence estimation} is essential for reliable predictions~\cite{gawlikowski2023survey,nair2020exploring, eggenreich2020variational,loquercio2020general, choi2019gaussian,han2022learning,han2023dynamic,yue2024deer}.
Recent work on large models focuses on estimating the confidence of LM-generated text through information-based~\cite{fomicheva2020unsupervised, kuhn2023semantic}, ensemble-based~\cite{malinin2020uncertainty}, density-based~\cite{yoo2022detection, ren2022out}, and reflexivity-based~\cite{kadavath2022language} methods. High-uncertainty content is often reviewed or processed by more advanced models for increased reliability~\cite{gupta2024language, yue2023large}. We recommend~\cite{fadeeva2023lm, geng2024survey} for a more comprehensive review. 

Our SEE uses the predictions of the small VLM by measuring its confidence to determine when to activate the larger VLM, enhancing the trade-off between performance and efficiency. In addition, a consistency criterion is proposed to facilitate the early-exiting decision making procedure.

\section{Small Guides Large (SGL) in VLMs}
We first provide the preliminaries of VLMs in Section~\ref{sec:preliminay}. Then, the proposed \textbf{training-free} Small VLM-Guided visual token Pruning (SGP) and Small VLM Early Exiting (SEE) techniques are detailed in Sections~\ref{sec:pruning} and \ref{sec:confidence}, respectively. 

\subsection{Preliminary of Vision-language Models} \label{sec:preliminay} 
VLMs~\cite{zhu2023minigpt, liu2023improvedllava, liu2023llava, chen2024far, wang2024qwen2} primarily follow a framework where a vision encoder~\cite{dosovitskiy2020image, radford2021learning, sun2023eva} converts an image into a sequence of visual tokens. These tokens are then combined with textual prompt tokens and fed into a language model~\cite{touvron2023llama, jiang2023mistral, team2023internlm, yang2024qwen2} to generate responses.

Specifically, an input image $\mathbf{I}$ is encoded into visual tokens by a vision encoder model $\text{VM}$:
\begin{equation}
    \mathbf{x}_{\text{I}} = \text{VM} (\mathbf{I}) \in \mathbb{R}^{N_{\text{I}} \times C},
\end{equation}
where $N_{\text{I}}$ represents the image token number.
The associated textual prompts are tokenized into prompt tokens $\mathbf{x}_{\text{T}} \in \mathbb{R}^{N_{\text{T}} \times C}$. As observed in ~\cite{chen2024image}, we usually have ${N_{\text{I}}} \gg {N_{\text{T}}}$, especially for high-resolution images.
The tokens $\mathbf{x}_{\text{I}}$ and $\mathbf{x}_{\text{T}}$ are concatenated and fed into a language model $\text{LM}$ to generate responses auto-regressively:
\begin{equation}
    \mathbf{p}_{\text{G}}^i = \text{LM} (\mathbf{x}_{\text{I}}, \mathbf{x}_{\text{T}}, \mathbf{x}_{\text{G}}^{1:i-1}) \in \mathbb{R}^{C_{\text{T}}},
\end{equation}
where $\mathbf{p}_{\text{G}}^i$ denotes the probability distribution 
over the vocabulary of size $C_{\text{T}}$.  The previous generated tokens $\mathbf{x}_{\text{G}}^{1:i-1}$ are used to predict the next token. The probability $\mathbf{p}_{\text{G}}^i$ is converted into the token embedding $\mathbf{x}_{\text{G}}^{i}$ via sampling, such as the argmax operation.

\begin{figure*}[!t]
    \centering
    \includegraphics[width=1\linewidth]{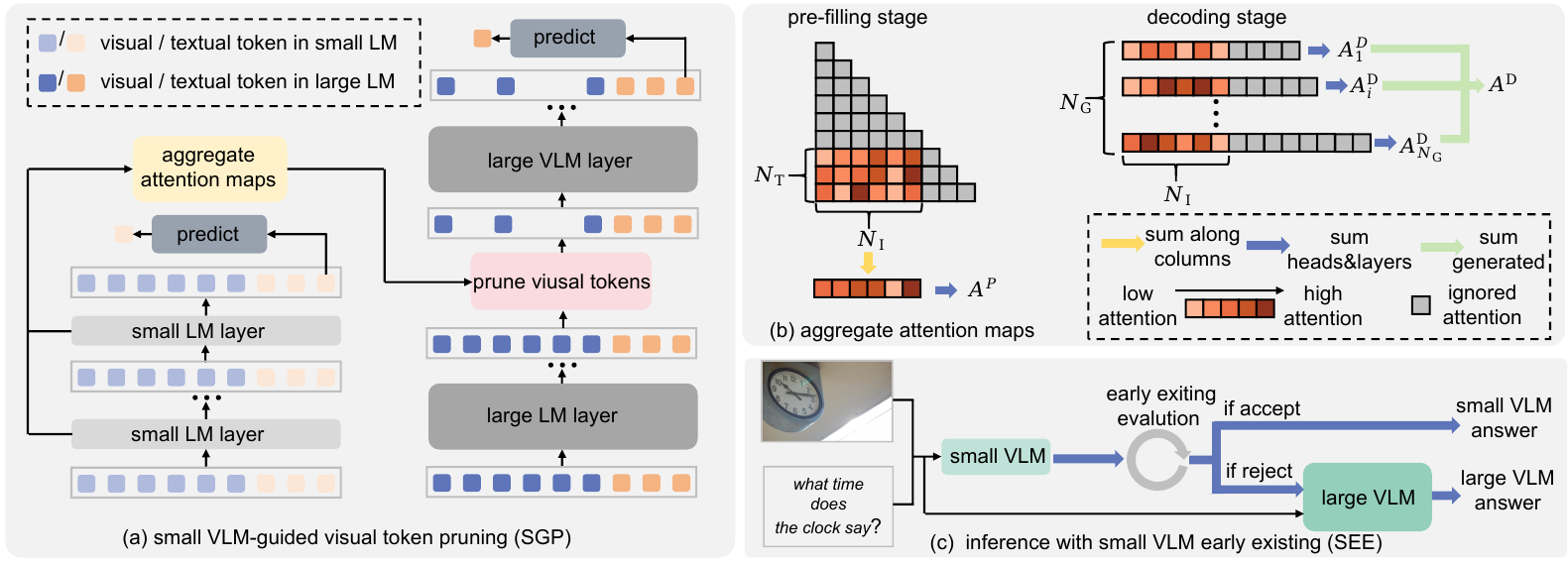}
    \vskip -0.1in
    \caption{Overview of SGL. \textbf{(a) Small VLM-guided visual token pruning in a large VLM (SGP).} We update a global attention map aggregated from all layer of a small VLM. This global attention map is used to rank visual tokens and guide the visual token pruning in a large VLM. \textbf{(b) Aggregation of attention maps in SGP.} We aggregate the attention score of visual tokens received from prompt tokens and generated tokens across all heads and layers in the small LM. Higher scores indicate greater significance. \textbf{(c) Inference with Small VLM Early Exiting (SEE)}. When the early exiting decision score from the small VLM is sufficient, the larger VLM will not be invoked. }
    \label{figure2}
    \vspace{-3mm}
\end{figure*}

\subsection{Small VLM-Guided Visual Token Pruning} \label{sec:pruning}
The inference efficiency of VLMs is greatly impacted by the large number of vision tokens.
A promising approach to mitigate this involves pruning less essential visual tokens using attention maps. However, pruning based on a single-layer attention map, as in~\cite{chen2024image}, falls short compared to using an \emph{oracle attention map aggregated from all layers} (Figure~\ref{figure1} (a)). 
Yet, obtaining this oracle attention map requires a full, computationally costly inference pass, making it impractical for real-world use. The key challenge is thus:

\emph{How can we efficiently acquire a precise attention map for effective visual token pruning?}

Based on our findings in Figure\ref{figure1}(b) that the aggregated attention map closely resemble that of a large VLM, we introduce SGP (Figure~\ref{figure2}(a)): using \emph{the small VLM's aggregated attention map} as an efficient and precise proxy to guide visual token pruning in a large VLM.

\paragraph{Aggregating attention maps in the small VLM.}
We initiate inference with a compact vision-language model, $\text{VLM}^{\text{S}}$ (\eg InternVL2-2B~\cite{chen2024far}), comprising a small vision model, $\text{VM}^{\text{S}}$, and a small language model, $\text{LM}^{\text{S}}$. 
This reduced model size significantly cuts computational costs compared to larger VLMs.
We input visual tokens $\mathbf{x}_{\text{I}} \in \mathbb{R}^{N_{\text{I}} \times C}$ from vision model $\text{VM}^{\text{S}}$ and textual prompt tokens $\mathbf{x}_{\text{T}} \in \mathbb{R}^{N_{\text{T}} \times C}$ into $\text{LM}_{\text{S}}$ to generate answers $\textbf{x}_{\text{G}} \in \mathbb{R}^{N_{\text{G}} \times C}$, where $N_{\text{G}}$ denotes the number of generated tokens.

The inference process of $\text{LM}_{\text{S}}$ involves a pre-filling stage followed by a decoding stage. We update two attention maps $\mathbf{A}^{\text{P}}, \mathbf{A}^{\text{D}}$ for the two stages, respectively.

\emph{\textcolor{blue}{(i)}  Pre-filling}. In this stage, attention maps are extracted from each layer and head, denoted as $\mathbf{A}_{j, k}^{\text{P}} \in \mathbb{R}^{(N_{\text{I}}+N_{\text{T}}) \times (N_{\text{I}}+N_{\text{T}})}$ , 
where $j$ and $k$ denote the layer and head index, respectively. Due to the causal nature of attention in $\text{LM}_{\text{S}}$, $\mathbf{A}_{j, k}^{\text{P}}$ is a lower triangular matrix. We specifically focus on the attention scores that visual tokens receive from prompt tokens. Therefore, we retrieve the bottom-left block
\begin{equation}
\mathbf{A}_{j, k}^{\text{P}} \in \mathbb{R}^{(N_{\text{I}}+N_{\text{T}}) \times (N_{\text{I}}+N_{\text{T}})}\Rightarrow  \tilde{\mathbf{A}}_{j, k}^{{\text{P}}} \in \mathbb{R}^{N_{\text{T}} \times N_{\text{I}}}.
\end{equation}

We then sum up the $N_\text{T}$ scores for each image token in $\tilde{\mathbf{A}}_{j, k}^{{\text{P}}}$, producing $\bar{\mathbf{A}}_{j, k}^{\text{P}} \in \mathbb{R}^{N_{\text{I}}}$.
During the forward pass, these attention maps are aggregated across layers and heads:
\begin{equation}
\mathbf{A}^{\text{P}} =\sum_{j=1}^{L} \sum_{k=1}^{H} \bar{\mathbf{A}}_{j, k}^{\text{P}}, 
\end{equation}
where $L$ and $H$ denote the number of layers and heads in $\text{LM}_{\text{S}}$, respectively. Note that we progressively update $\mathbf{A}^{\text{P}}$ in an \emph{accumulative} manner without cacheing all $\mathbf{A}_{j, k}^{\text{P}}$.
This procedure is illustrated in Figure~\ref{figure2}(b), left.

\emph{\textcolor{blue}{(ii)} Decoding}. Attention scores of $N_{\text{I}}$ visual tokens from the $i$-th generated token can be denoted as $\mathbf{A}^{\text{D}}_{i,j,k} \in \mathbb{R}^{N_{\text{I}}}$ for head-$k$ in layer-$j$. These scores are accumulated as
\begin{equation}
\mathbf{A}^{\text{D}} =\sum_{i=1}^{N_{\text{G}}} \sum_{j=1}^{L} \sum_{k=1}^{H} \mathbf{A}_{i, j, k}^{\text{D}}.
\end{equation}
The aggregation in the decoding phase is visualized in Figure~\ref{figure2}(b), right.
Upon inference completion, we calculate the overall attention scores for vision tokens via $\mathbf{A} = \mathbf{A}^{\text{P}}+\mathbf{A}^{\text{D}}$. This comprehensive assessment $\mathbf{A}$ is further employed to rank and prune visual tokens.

\paragraph{Visual token pruning in the large VLM.}
To improve the efficiency of a larger model $\text{VLM}^{\text{L}}$ (\eg InternVL2-26B~\cite{chen2024far}), we prune less important visual tokens based on the ranking obtained from $\mathbf{A}$.  Specifically, the same image is fed into its vision model $\text{VM}^{\text{L}}$, producing visual tokens. Since $\text{VLM}^{\text{S}}$ and $\text{VLM}^{\text{L}}$ share the same architecture suite, $\text{VM}^{\text{L}}$ outputs the same number of visual tokens. These tokens, combined with prompt tokens, are fed into $\text{LM}^{\text{L}}$. Inspired by FastV~\cite{chen2024image}, we retain only the top $R$\% of important visual tokens in an intermediate layer of $\text{LM}^{\text{L}}$, as determined by the ranking. Owing to the comprehensive importance score from the small VLM, we can apply a low retention ratio (\eg 5\%) at an early layer (\eg, the $2$-rd layer), significantly reducing the computational cost of  $\text{VM}^{\text{L}}$.

\subsection{Small VLM Early Exiting} \label{sec:confidence}
While SGL effectively reduces the token load in the large VLM, incorporating a small VLM does add some overhead relative to using the large VLM alone. Fortunately, the performance gap between small and large VLMs is relatively minor compared to their computational difference, indicating that the small VLM’s outputs are often quite competitive. This inspires us to devise Small VLM Early Exiting (SEE), maximizing the utility of the small VLM by assessing its outputs. For some ``easy'' questions, the inference pipeline exits early after obtaining the small VLM's prediction, further enhancing the inference efficiency. We demonstrate its pipeline in Figure~\ref{figure2} (c).

During small VLM inference, the token generation probability can be recorded to estimate the answer confidence. A straightforward yet effective method for confidence estimation involves calculating the length-normalized sequence probability~\cite{murray2018correcting, fadeeva2023lm}, which can be expressed as:
\begin{equation}\label{eq:confidence}
  \mathcal{S}_{\text{confidence}}  = \exp \left\{\frac{1}{{N_{\text{G}}}} \log P(\mathbf{x}_{\text{G}}^1, ... \mathbf{x}_{\text{G}}^{N_{\text{G}}}) \right\}, 
\end{equation}
where
\begin{equation}
P(\mathbf{x}_{\text{G}}^1, ... \mathbf{x}_{\text{G}}^{N_{\text{G}}}) = \prod_{i=1}^{N_{\text{G}}} P \left(\mathbf{x}_{\text{G}}^i \mid \text{LM}^{\text{S}} (\mathbf{x}_{\text{I}}, \mathbf{x}_{\text{T}}, \mathbf{x}_{\text{G}}^{1:i-1}) \right).
\end{equation}

\begin{table*}[t]
\centering

\tablestyle{4.6pt}{1.2}
\begin{tabular}{c c  |c c c c|  c c c c | c c c| c }

    \multirow{2}{*}{method} & \multicolumn{1}{c|}{\multirow{1}{*}{ token}}   & \multicolumn{4}{c|}{visual question answering}  & \multicolumn{4}{c|}{comprehensive benchmark}  & \multicolumn{3}{c|}{visual grounding}   & \multicolumn{1}{c}{score} \\

    \multirow{1}{*}{} & \multicolumn{1}{c|}{\multirow{1}{*}{ratio}}   & \multicolumn{1}{c}{TextVQA}  & \multicolumn{1}{c}{ChartQA}  & \multicolumn{1}{c}{DocVQA} & \multicolumn{1}{c|}{GQA} & \multicolumn{1}{c}{SEED} & \multicolumn{1}{c}{MMBench} & \multicolumn{1}{c}{MM-Vet} & \multicolumn{1}{c|}{MME}  & \multicolumn{1}{c}{RC}  & \multicolumn{1}{c}{RC+}  & \multicolumn{1}{c|}{RC-g}  & \multicolumn{1}{c}{ratio} \\
  \midrule[1.2pt]
  
   InternVL2-26B~\cite{chen2024far} & 100\%  & 82.45 & 84.92 & 92.14 & 64.89 & 76.78 & 83.46 & 64.00 & 2270 & 91.24 & 86.67 & 88.44 & 100.00\% \\     

   InternVL2-2B~\cite{chen2024far} & 100\%  & 73.19 & 76.24 & 85.93 & 61.16 & 71.62 & 72.93 & 43.30 & 1878 & 82.26 & 73.53  & 77.55 & 87.25\% \\     \hshline

  \multirow{3}{*}{26B w/ ToMe~\cite{bolya2022token}} & 64\%  & 80.22  & 76.24 & 79.51 & 64.49 & 75.60 & 82.74 & 60.10 & 2235  & 84.02 & 78.91 & 80.35 & 94.24\%\\

  {}  & 35\%  & 75.74 & 62.44   & 66.79 & 63.61 & 73.84 & 81.28 & 52.50 & 2178 & 71.08 & 64.97 & 68.08 & 85.20\% \\

  {} & 9\% & 51.69 & 28.60 & 28.46 & 57.52 & 65.19 & 73.09 & 37.70 & 1933 & 20.33 & 17.74 & 19.36 & 54.28\% \\  \hshline

\multirow{3}{*}{ 26B w/ FastV~\cite{chen2024image}} & 64\%  & 82.26 & 85.08  & 92.20 & 64.80 & 76.81 & 83.24 & 63.20 & 2270 & 91.30 & 86.66 & 88.30 & 99.84\%\\
 {} & 35\%  & 75.62 & 71.68 & 68.32 & 61.20 & 71.64 & 78.31 & 45.00 & 2140 & 85.06 & 77.61 & 81.39 & 88.28\% \\
 {} & 9\%  & 43.84 & 26.20 & 26.81 & 44.90 & 54.56 & 62.33 & 31.60 & 1799 & 19.65  & 16.66 & 17.22 & 46.99\%\\ \hshline

\multirow{3}{*}{ 26B w/ SGP (ours)} & 64\%  & 82.41 & 85.04 & 92.12 & 65.07 & 76.71 & 83.30 & 65.60 & 2259 & 91.07 & 86.71 & 88.05  & 100.14\%\\
    {} & 35\%  & 81.97 & 81.68 & 91.14 & 64.62 & 75.72 & 82.17 & 63.20 & 2258 & 89.38 & 84.35 & 86.07 & 98.36\%\\
    {} & 9\%  & 78.98  & 72.96  & 87.26 & 62.10 &  72.23 & 75.56 & 52.10 & 2004 & 80.36 & 72.22 & 77.45 & 89.58\% \\

    \end{tabular}
    \vskip -0.1in
\caption{\textbf{Comparison between SGP and previous visual token pruning methods.} ``Token ratio'' denotes the average ratio of retrained visual tokens. ``26B'' denotes the original InternVL2-26B. In ``26B w/ SGP (ours) '', we employ the aggregated attention map across all layers in InternVL2-2B to guide the visual token pruning in  InternVL2-26B. For fair comparison, we do not employ SEE in these experiments. The ``\textbf{score ratio}'' is obtained by calculating the ratio of each score relative to InternVL2-26B, followed by averaging these ratios.} %
    \label{tab:SOTA_ratio}
    \vspace{-3mm}
\end{table*}

In our token pruning scenario, apart from the naive confidence metric $\mathcal{S}_{\text{confidence}}$, we further propose a \textbf{consistency score} for making early-exiting decisions. Specifically, we can naturally hypothesize that the small VLM accurately identifies essential visual tokens when it provides a correct answer. Conversely, if the small VLM's answer is correct, a consistent prediction should be obtained when visual tokens are pruned by SGP (Section~\ref{sec:pruning}). In this basis, we introduce a consistency score $\mathcal{S}_{\text{consistency}}$ to measure the consistency of the generation after visual token pruning. A higher score indicates a higher probability that the small VLM yields a correct answer, where early exiting is more reliable. Let $\text{LM}^{\text{S}^{\prime}}$ represent the small language model with pruned visual tokens, the consistency score is obtained by
\begin{equation} \label{eq:consis}
  \mathcal{S}_{\text{consistency}} = \prod_{i=1}^{N_{\text{G}}} P \left(\mathbf{x}_{\text{G}}^i \mid \text{LM}^{\text{S}^{\prime}} (\mathbf{x}_{\text{I}}, \mathbf{x}_{\text{T}}, \mathbf{x}_{\text{G}}^{1:i-1}) \right).
\end{equation}

It is worth noting that the calculation of $\mathcal{S}_{\text{consistency}}$ is extremely efficient, because: \textcolor{blue}{(\emph{i})} in Equation~\ref{eq:consis}, visual tokens $\mathbf{x}_{\text{I}}$, text tokens $\mathbf{x}_{\text{T}}$ and the generated tokens $\mathbf{x}_{\text{G}}$ have all been obtained in Section~\ref{sec:pruning}, thus the consistency score can be computed in parallel rather than autoregressively; \textcolor{blue}{(\emph{ii})} a high pruning ratio significantly reduces computational cost. We find that removing 95\% visual tokens is feasible in practice. In this scenario, calculating $\mathcal{S}_{\text{consistency}}$ requires \textless10\% of the initial inference time with the small VLM. Finally, we compute the final early-exiting \textbf{decision score}:

\begin{equation}
    \mathcal{S} = \frac{1}{2}(\mathcal{S}_{\text{confidence}} + \mathcal{S}_{\text{consistency}}).
\end{equation}
The inference pipeline exits early at the small VLM when the score is above a predefined threshold. In Section~\ref{sec:exp}, we empirically show that our early-exting criterion $\mathcal{S}$ outperforms other early-exiting criteria, such as quantile~\cite{gupta2024language}, entropy~\cite{fadeeva2023lm}, and either one item $\mathcal{S}_{\text{consistency}}$ or $\mathcal{S}_{\text{confidence}}$.


To sum up, the small VLM in our SGL plays two roles. For any input, it first performs inference, producing 

\textcolor{blue}{(\emph{i})} The importance scores $\mathbf{A}$ for vision tokens (SGP). 

\textcolor{blue}{(\emph{ii})} An early prediction and the corresponding early-exiting decision score $\mathcal{S}$ (SEE);

If the large VLM is decided to be activated by SEE, SGP prunes a large amount of unimportant visual tokens to accelerate the inference of the large VLM.

\section{Experiment} \label{sec:exp}

\subsection{Experimental Setup}

\paragraph{Models.}
We conduct experiments using InternVL2~\cite{chen2024far}, which provides checkpoints for various model sizes, facilitating the exploration of small VLMs in guiding visual token pruning in large VLMs. Specifically, we use InternVL2-2B as the small VLM and InternV2L-26B as the large VLM dy default. We also include Qwen-VL~\cite{wang2024qwen2} and LLaVa-OV~\cite{li2024llava} to evaluate the generalizability of our method. The default setting in experiments is marked in \colorbox{bestcolor}{color}.

\begin{figure*}[!t]
    \centering
    \includegraphics[width=1\linewidth]{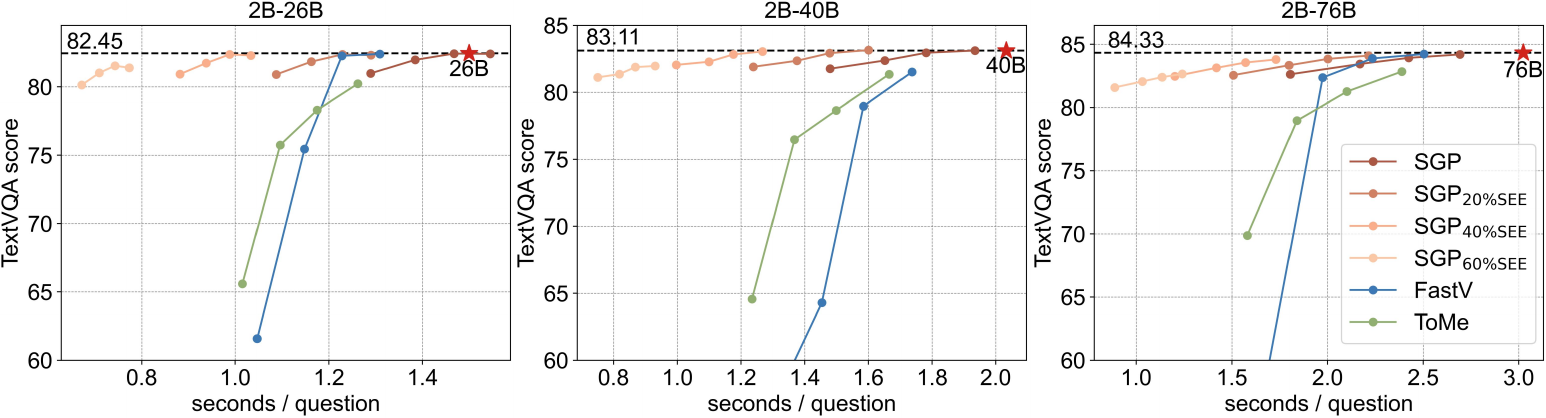}
    \vskip -0.1in
    \caption{\textbf{Performance-efficiency curves of SGL (SGP + SEE).}  The results with 18\%, 35\%, 50\%, and 64\% visual token retention ratios are presented as a curve. For the 26B and 40B, we use an NVIDIA H20 GPU, and the 76B  is sharded on two GPUs.} 
    \label{fig:trade_off}
    \vspace{-4mm}
\end{figure*}

\paragraph{Evaluation benchmarks.}
We conduct experiments on four VQA benchmarks including TextVQA~\cite{singh2019towards}, ChartQA~\cite{masry2022chartqa}, DocVQA~\cite{mathew2021docvqa}, and GQA~\cite{hudson2019gqa}.  To evaluate performance in visual grounding, we introduce RefCOCO (RC)~\cite{yu2016modeling}, RefCOCO+ (RC+)~\cite{yu2016modeling}, and RefCOCOg (RC-g)~\cite{mao2016generation}. Additionally, we assess the model's capability in general multi-modal understanding on comprehensive benchmarks such as SEED\cite{li2023seed}, MMBench~\cite{liu2025mmbench}, MM-Vet~\cite{yu2023mm}, and MME~\cite{fu2023mme}. 

\subsection{Comparing SGP with Previous Methods}
We first validate the effectiveness of our SGP without the early-exiting mechanism. The comparison with representative visual token compression methods, including ToMe~\cite{bolya2022token} and FastV~\cite{chen2024image}, is presented in Table~\ref{tab:SOTA_ratio}, across different average visual token retention ratios.  All experiments are conducted based on InternVL2-26B model, consisting of 48 layers. For our method and FastV~\cite{chen2024image}, we prune 60\%, 80\% and 95\% visual tokens at the 19-th, 9-th, and 2-th layer, achieving average token retention ratios of 64\%, 35\%, and 9\%, respectively. ToMe~\cite{bolya2022token} performs token merging prior to the language model, with the merging ratio adjusted to achieve similar average token retention ratios.

At a relatively high token retention ratio, such as 64\%, all methods exhibit competitive performance across various tasks. This suggests significant visual token redundancy in VLMs, underscoring the importance of visual token pruning.

When the token retention ratio is decreased to 35\%, the performance of ToMe and FastV starts to drop, particularly in OCR-related tasks, including TextVQA~\cite{singh2019towards}, ChartQA~\cite{masry2022chartqa}, and DocVQA~\cite{mathew2021docvqa} as well as visual grounding tasks. Their performance on MM-Vet~\cite{yu2023mm} also drops significantly, since it also includes many OCR-related questions. These tasks require methods to accurately retain answer-related visual tokens to understand image details. This performance decline demonstrates that ToMe and FastV can not accurately retain essential tokens. In contrast, our method maintains competitive performance across all tasks.

With only 9\% of visual tokens retained, FastV and ToMe collapse across all tasks, as critical visual tokens are lost due to inaccurate pruning. In this challenging scenario, our method experiences only a marginal performance drop compared to other methods, achieving over 89\% of the original InternVL2-26B's performance. The visualization in Figure~\ref{fig:visualization} also validates the supeority of our SGP, which successfully preserves the tokens most relevant to a correct answer, thanks to the global attention maps aggregated from the small VLM.

\subsection{SGP with SEE Towards Improved Efficiency}
We further validate the supeority of our SGL by incorporating both SGP and SEE mechanisms. 
The performance-efficiency curves for varying token retention ratios (18\%, 35\%, 50\%, and 64\%) across large VLMs of different sizes (InternVL2-\{26B, 40B, 76B\}) on TextVQA are presented in Figure~\ref{fig:trade_off}. As discussed in Section~\ref{sec:confidence}, we perform early existing based the decision score of the small VLM's answers to reduce the invocation of the large VLM. When the large VLM is activated, SGP is employed to reduce the visual token redundancy. Note that the ratio of early exiting can be flexibly controlled by adjusting the decision threshold, as a smaller threshold induces a lower ratio, \emph{i.e.} less invocation for the large VLM. Here we present the performance curves at 60\%, 40\%, and 20\% early exiting ratios, denoted as $\text{SGP}_{\text{60\%SEE}}$, $\text{SGP}_{\text{40\%SEE}}$, and $\text{SGP}_{\text{20\%SEE}}$, respectively.

It can be observed that, with the 26B large VLM, our method SGP without SEE yields slower inference compared to FastV and ToMe, due to the overhead of the 2B small VLM. However, scaling the large VLM to 40B and 76B results in competitive inference speeds and superior performance relative to FastV and ToMe, particularly at low token retention ratios. Additionally, the proposed SEE enables SGP to maintain competitive performance at 20\% and 40\% early-exiting ratios while significantly reducing the average inference time across all VLM sizes. These results demonstrate the effectiveness of our SEE in identifying unreliable answers from the small VLM and appropriately invoking the large VLM. To summarize, our SGL offers superior trade-off between efficiency and performance.

\begin{table}[t]  
\renewcommand{\arraystretch}{1.0}
    \centering

\tablestyle{4.6pt}{1.2}

  \begin{tabular}{c |  c c c| c }

    \multirow{1}{*}{attention map source} & \multirow{1}{*}{TextVQA} & \multirow{1}{*}{SEED} & \multirow{1}{*}{RC}  &{score ratio} \\
  \midrule[1.2pt]
all layers of 26B (oracle) & 80.04 & 71.90 & 84.49 &  94.44\% \\
\hshline
one layer of 26B (FastV \cite{chen2024image}) & 43.84 & 54.56 & 20.33  & 48.37\%  \\
\hshline
10\% layers of 2B  & 44.40 & 63.03 & 18.09  & 51.92\%\\
30\% layers of 2B  & 57.42 & 63.07 & 15.20  & 56.15\%\\
50\% layers of 2B  & 74.16 & 68.17 & 56.74  & 80.31\% \\
70\% layers of 2B & 77.29 & 70.96 & 80.33  & 91.40\%\\
\hshline
\bestcell{all layers of 2B (ours)} & \bestcell{78.98} & \bestcell{72.23} & \bestcell{80.36}  & \bestcell{92.64\%}\\
    
    \end{tabular}
    \vskip -0.1in
    \caption{\textbf{Performance with attention maps from different sources.} The visual token retention ratio is set to 9\% for all experiments. Aggregating attention maps from all layers of the small model (2B) achieves performance comparable to the oracle.}
    \vspace{-3mm}
\label{tab:abl_source}
\end{table}

\subsection{Ablation Study of Key Designs}

\vspace{-1mm}
\paragraph{Effectiveness of all-layer attention maps.}
To verify the superiority of aggregating attention maps from all layers of the small VLM, we experiment using different sources to guide visual token pruning. All experiments are conducted with a 9\% retention ratio without SEE. 

The results shown in Table~\ref{tab:abl_source} indicate that using attention maps aggregated from the small model outperforms using a single layer from the large model, such as FastV \cite{chen2024image}. Moreover, the average performance consistently improves when the number of aggregated layers increases. This demonstrates that leveraging attention maps from multiple layers helps accurately retain essential visual tokens, achieving performance comparable to the oracle. Notably, our method even slightly outperforms the oracle on the SEED benchmark, highlighting its effectiveness.

\vspace{-1mm}
\paragraph{Key tokens used for attention aggregation.}
Our SGP adopts both prompt and generated tokens in the attention maps to assess the importance of visual tokens. We ablate this design choice without SEE in Table~\ref{tab:abl_token_used}. It can be observed that using only generated or prompt tokens might result in unstable performance. For example, generated tokens achieve the best performance on the RC dataset but perform poorly on the SEED benchmark. Similarly, using only the last prompt token, as in FastV \cite{chen2024image}, also leads to instability, particularly on the RC dataset. These results demonstrate that employing both prompt and generated tokens provide a comprehensive evaluation of visual tokens.

\begin{table}[t]  
\renewcommand{\arraystretch}{1.0}
    \centering

\tablestyle{4.6pt}{1.2}

  \begin{tabular}{c |  c c c| c }

    \multirow{1}{*}{used token} & \multirow{1}{*}{TextVQA} & \multirow{1}{*}{SEED} & \multirow{1}{*}{RC}  &{score ratio} \\
  \midrule[1.2pt]
  {last prompt token}  & {79.40} & {69.53} & {13.63} & 67.26\% \\
 {prompt tokens}  & {76.15} & {72.25} & {59.90} & 84.03\%\\
 {generated tokens} & {79.38} & {63.47} & {83.51}  & 90.16\% \\
 \bestcell{prompt + generated tokens}  & \bestcell{78.98} & \bestcell{72.23} & \bestcell{80.36}  & \bestcell{92.64\%} \\

    \end{tabular}
    \vskip -0.1in
    \caption{\textbf{Performance of using different tokens in visual token importance evaluation.}  Prompt and generated tokens provide a comprehensive evaluation of visual tokens }
\label{tab:abl_token_used}
\vskip -0.15in
\end{table}

\begin{figure}[!t]
    \centering
    \includegraphics[width=0.9\linewidth]{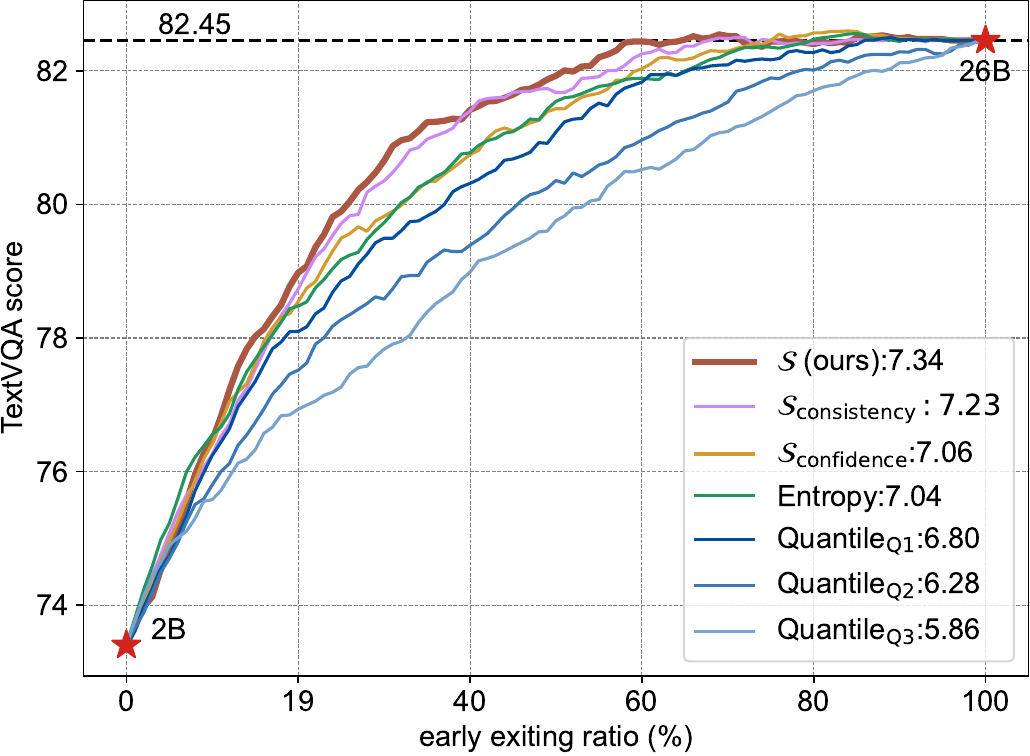}
    \vskip -0.1in
    \caption{\textbf{Comparison of different early-exiting decision scores.} We present the area between each strategy's curve and the 2B model score alongside their names. A larger area indicates a more effective criterion. With the same early exiting ratio, a higher score reflects improved accuracy in identifying incorrect responses from the small VLM. Note that SGP is not adopted for clear comparison.} 
    \label{fig:confidence_compare}
    \vspace{-5mm}
\end{figure}

\paragraph{Different strategies used in SEE.} We further investigate the effectiveness of the early-exiting criteria used in our SEE. The proposed strategy $\mathcal{S}$ is compared with other strategies, including only length-normalized sequence probability ($\mathcal{S}_{\text{confidence}}$ in Equation~\ref{eq:confidence})~\cite{murray2018correcting, fadeeva2023lm}, consistency score ($\mathcal{S}_{\text{consistency}}$ in Equation~\ref{eq:consis}), quantile~\cite{gupta2024language} and entropy~\cite{fadeeva2023lm}.
For the quantile strategy, the top 75th, 50th, and 25th percentile probabilities among generated tokens are used as confidence scores, denoted as $\text{Quantile}_{\text{Q1}}$,  $\text{Quantile}_{\text{Q2}}$, and  $\text{Quantile}_{\text{Q3}}$, respectively. In the entropy strategy, we aggregate the entropy of each generated token, where higher entropy indicates lower confidence. The results are presented in Figure~\ref{fig:confidence_compare}, where SGP is omitted in all experiments.

It is observed that $\mathcal{S}_{\text{confidence}}$ outperforms other baselines except for $\mathcal{S}_{\text{consistency}}$. Building on this, we develop our strategy $\mathcal{S}$ by integrating both $\mathcal{S}_{\text{confidence}}$ and $\mathcal{S}_{\text{consistency}}$, achieving the outstanding performance. It is noteworthy that the calculation of $\mathcal{S}_{\text{consistency}}$ is efficient, consuming $<$1/10 of the initial small VLM inference time, as analysed in Section~\ref{sec:confidence}.

\begin{figure*}[!t]
    \centering
    \includegraphics[width=1\linewidth]{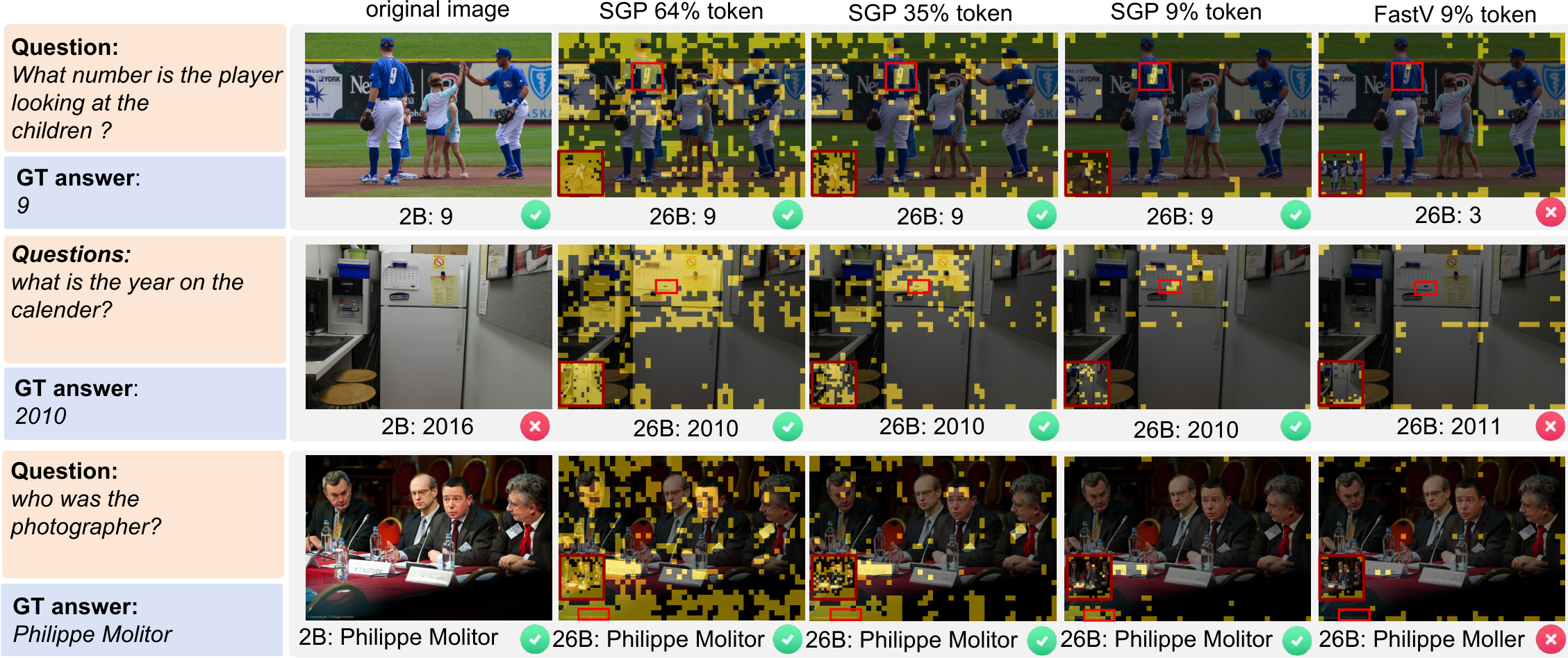}
    \vskip -0.1in
    \caption{\textbf{Visualization of SGP under different visual token retention ratios and answers.}  Visual tokens are pruned by 60\%, 80\%, and 95\% at the 19th, 9th, and 2nd layers of the large VLM of 26B, which comprises 48 layers. This results in average token retention ratios of 64\%, 35\%, and 9\%, respectively.     Retained tokens are highlighted with \raisebox{0.25em}{\colorbox{important_token!50}{}}.  Thumbnails employed in InternVL are presented in the left corner. } 
    \label{fig:visualization}
    \vskip -0.15in
\end{figure*}

\subsection{Visualization of Token Pruning and Answers}
To provide a deeper understanding of our SGP, we visualize the token pruning results in Figure~\ref{fig:visualization}. The first row demonstrates that the small VLM helps retain essential visual tokens necessary for answering questions across various token retention ratios. The second row shows that, even when the answer is incorrect, the small VLM can still retain tokens related to the answer. This suggests that, although the small VLM lacks the precise reasoning and perception necessary for accurate answers, it possesses sufficient reasoning ability to identify target regions, thereby guiding visual token pruning in the large VLM.
In the last row, we present a challenging example where the answer is located at the boundary of the image and appears in a small font size. In this difficult scenario, the small VLM produces a correct answer as the large VLM, verifying competitive performance of the small VLM. This enable us to conduct SEE using the responses from the small VLM during inference.

We also present the visual tokens pruning and answers from FastV~\cite{chen2024image}. We find that it can only preserve partial tokens relevant to the answer in the thumbnail and fails to preciously retain them in the main image.  Consequently, this limitation impairs the model's ability to perceive image details, leading to inaccurate predictions.

\subsection{Generalization on Different Size Models}

\paragraph{Small VLM.}
In Table~\ref{tab:abl_small_size}, we evaluate the effectiveness of using small VLMs of varying sizes (InternVL2-\{1B, 2B, 4B\}) to guide visual token pruning in the larger InternVL2-26B model~\cite{chen2024far}. These models are constructed using different language models (LMs) from various sources. Specifically, InternVL2-1B uses Qwen2~\cite{yang2024qwen2}, while InternVL2-2B and InternVL2-4B adpot InterLM2~\cite{cai2024internlm2} and Phi3~\cite{abdin2024phi}, respectively. The results show that our method is robust to the choice of small model and maintains compatibility with different LMs. Surprisingly, InternVL2-1B performs slightly better than InternVL2-2B across three tasks, motivating further reducing the small VLM size in future studies.

\begin{table}[t]  
\renewcommand{\arraystretch}{1.0}
    \centering

\tablestyle{4.6pt}{1.2}

  \begin{tabular}{c | c | c c c| c }

    \multirow{2}{*}{small VLM} & \multirow{2}{*}{LM source}  & \multirow{2}{*}{TextVQA} & \multirow{2}{*}{SEED} & \multirow{2}{*}{RC}  &\multirow{2}{*}{score ratio} \\

  {size} & {} & & & & \\
    \midrule[1.2pt]
1B & Qwen2~\cite{yang2024qwen2} & 79.38 & 72.27 & 82.06  & 93.44\% \\

\bestcell{2B} & \bestcell{InterLM2~\cite{cai2024internlm2}}  & \bestcell{78.98} & \bestcell{72.23} & \bestcell{80.36}  & \bestcell{92.64\%} \\

4B & Phi3~\cite{abdin2024phi} & 79.70 & 73.65  & 75.39   & 91.73\% \\
    
    \end{tabular}
    \vskip -0.1in
    \caption{\textbf{Performance of leveraging  small VLMs of different sizes.}  These small VLMs are based on various language models.  }
        \vspace{-3mm}
\label{tab:abl_small_size}
\end{table}


\paragraph{Large VLM.} We further substitute the large VLM in SGL by InternVL2-40B and InternVL2-76B, with the small VLM fixed as InternVL2-2B. The results in Table~\ref{tab:abl_large_size} indicate that our small model effectively guides significantly larger VLMs. This suggests that SGP is robust and has potential to guide the visual token pruning in huge VLMs.

\begin{table}[t]  
\renewcommand{\arraystretch}{1.0}
    \centering
\normalsize
\tablestyle{4.6pt}{1.2}

  \begin{tabular}{c | c | c c c| c }

    \multirow{2}{*}{large VLM} & \multirow{2}{*}{w/ ours}  & \multirow{2}{*}{TextVQA} & \multirow{2}{*}{SEED} & \multirow{2}{*}{RC}  &\multirow{2}{*}{score ratio} \\

  {size} & {} & & & & \\
    \midrule[1.2pt]
26B & \ding{56} & 82.45 & 76.78 & 91.24  & 100.00\% \\
\bestcell{26B} & \bestcell{\ding{52}} & \bestcell{78.98} & \bestcell{72.23} & \bestcell{80.36}  & \bestcell{92.64\%} \\ \hshline

40B & \ding{56} & 83.11 & 78.15  & 93.00  & 100.00\%\\
40B & \ding{52} & 79.96 & 74.11  & 79.99 & 92.38\% \\ \hshline

76B & \ding{56}  &  84.33 & 78.17  & 92.20  & 100.00\%\\
76B & \ding{52}  &  80.72 & 73.93  & 81.82  & 92.98\% \\
    
    \end{tabular}
    \vskip -0.1in
    \caption{\textbf{Visual token pruning for different-sized large VLMs.} The average retention ratio is set to 9\%.}
    \vspace{-4mm}
\label{tab:abl_large_size}
\end{table}


\subsection{Generalization on Various Architectures}
We further assess the generalizability of SGL using Qwen2-VL~\cite{wang2024qwen2} and LLaVa-OV~\cite{li2024llava}. 
The smallest models in the families, Qwen2-VL-2B and LLaVa-OV-0.5B, are used to guide visual token pruning in the largest models, Qwen2-VL-72B and LLaVa-OV-72B. The results in Table~\ref{tab:general} reveal that SGL enables the large VLM to maintain approximately 96\% of their original performance while achieving a retention ratio of 9\% for visual tokens. This underscores the potential applicability of our method to varied VLM architectures.

\begin{table}[t]  
\renewcommand{\arraystretch}{1.0}
    \centering

\tablestyle{4.6pt}{1.2}

  \begin{tabular}{c | c | c | c }

    \multirow{1}{*}{method} & \multirow{1}{*}{token ratio}  & \multirow{1}{*}{TextVQA}   &\multirow{1}{*}{score ratio} \\

    \midrule[1.2pt]
    Qwen2-VL-72B~\cite{wang2024qwen2} & 100\% & 85.50 & 100\% \\
    w/ SGP (ours) & 64\% & 85.49  & 99.98\% \\
    w/ SGP (ours) & 35\% & 85.13 & 99.56\% \\
    w/ SGP (ours) & 9\%  & 82.88 & 96.94\% \\
\hshline
    LLaVa-OV-72B~\cite{li2024llava} & 100\% & 79.30 & 100\% \\
    w/ SGP (ours) & 64\% &  79.19 & 99.86\% \\
    w/ SGP (ours) & 35\% & 78.65  & 99.18\% \\
    w/ SGP (ours) & 9\%  & 75.98  & 95.81\% \\

    \end{tabular}
    \vskip -0.1in
    \caption{\textbf{Generalizability on Qwen2-VL and LLaVa-OV.}  We adopt Qwen2-VL-2B and LLaVa-OV-0.5B to guide the visual token pruning in Qwen2-VL-72B and LLaVa-OV-0.5B, respectively.}
\label{tab:general}
\vspace{-5mm}
\end{table}

\section{Conclusion}
In this study, we explore the effectiveness of attention maps for visual token pruning in VLMs. Our findings reveal that the attention map aggregated from all layers of a small VLM exhibits patterns akin to that of a larger VLM. Based on this insight, we introduce SGP, which prunes visual token in large VLMs under the guidance from a small VLM. This small VLM is further exploited to perform early exiting (SEE) to make full use of its predictions. Both of these two techniques are training-free. Comprehensive experiments across 11 benchmarks demonstrate the method's effectiveness, particularly at low visual token retention ratios.

\paragraph{Limitations and future works.}
Our method is primarily validated on multi-modal understanding tasks. Its application in recent VLMs~\cite{sikder2023transfusion, xie2024showo, wu2024vila, wang2024emu3}, unifying both understanding and generation, is worth studying in the future.

{
    \small
    \bibliographystyle{ieeenat_fullname}
    \bibliography{main}
}
\cleardoublepage
 
\appendix
\onecolumn
\begin{center}
    \section*{Appendix}
\end{center}


\section{SGP with SEE Towards Improved Efficiency}
\begin{figure*}[!h]
    \centering
    \includegraphics[width=1\linewidth]{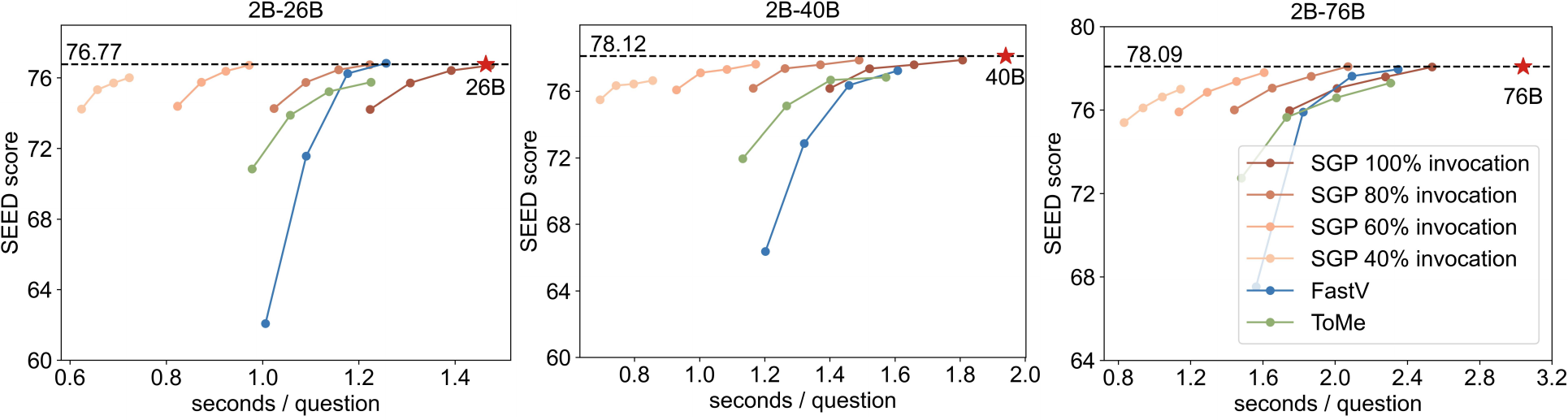}
    \caption{Performance-efficiency curves of SGL (SGP + SEE) on SEED~\cite{li2023seed}. The results with 18\%, 35\%, 50\%, and 64\% visual token retention ratios are presented as a curve. For the 26B and 40B, we use an NVIDIA H20 GPU, and the 76B is sharded on two GPUs. } 
    \label{app_fig:plot_seed}
\end{figure*}

\begin{figure*}[!h]
    \centering
    \includegraphics[width=1\linewidth]{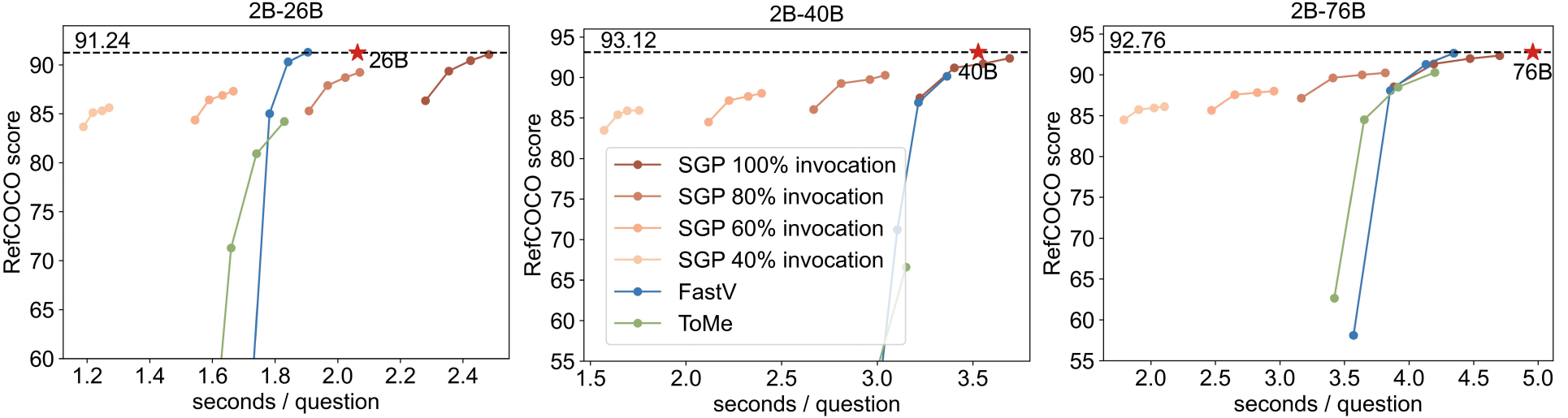}
    \caption{Performance-efficiency curves of SGL (SGP + SEE) on RefCOCO~\cite{yu2016modeling}. The results with 18\%, 35\%, 50\%, and 64\% visual token retention ratios are presented as a curve. For the 26B and 40B, we use an NVIDIA H20 GPU, and the 76B is sharded on two GPUs. } 
    \label{app_fig:plot_refcoco}
\end{figure*}

In Figure~\ref{app_fig:plot_seed} and Figure~\ref{app_fig:plot_refcoco}, we further validate the superiority of our SGL by incorporating both SGP and SEE mechanisms, on SEED~\cite{li2023seed} and RefCOCO~\cite{yu2016modeling} benchmarks. It can be observed that, with the 26B large VLM, our method SGP without SEE yields slower inference compared to FastV and ToMe. This is attributed to the computational overhead of the 2B small VLM, particularly on RefCOCO, where it requires a non-negligible amount of time to auto-regressively generate a greater number of tokens compared to other datasets \eg SEED. However, scaling the large VLM to 40B and 76B
results in competitive inference speeds and superior performance relative to FastV and ToMe, particularly at low token
retention ratios. 

\section{Memory Efficiency}
In this section, we analyze the memory allocation of SGL. Our approach incorporates a small VLM \eg InternVL2-2B in addition to the large VLM, which may introduce some additional memory overhead. Fortunately, the small VLM consumes only a minimal portion of memory compared to the large model. As a result, our method retains memory efficiency, as verified in Table~\ref{app:tab_memory}.

\begin{table*}[!h]  
\renewcommand{\arraystretch}{1.0}
    \centering

\tablestyle{4.6pt}{1.2}

  \begin{tabular}{c  c  c  cc c}

    \multirow{1}{*}{small VLM}  & \multirow{1}{*}{small VLM memory}  & \multirow{1}{*}{large VLM}  & \multirow{1}{*}{large VLM peak memory} & \multirow{1}{*}{large VLM with SGL peak memory} & \multirow{1}{*}{$\Delta$} \\
        \midrule[1.2pt]
    2B & 4.48 GiB & 26B & 51.60 GiB & 54.24 GiB  & +2.64GiB (5.11\%)\\

    2B & 4.48 GiB & 40B & 77.94 GiB &  80.60 GiB  & +2.66GiB (3.41\%) \\

    2B & 4.48 GiB & 76B & 147.64 GiB &  147.25 GiB & -0.39 GiB (0.26\%)  \\

    \end{tabular}
    \caption{\textbf{Mmeory analysis of SGL}. The meory of our method is measured with 9\% average retention ratio. ``small VLM memory'' refers to the memory required to load the single small VLM. ``Large VLM peak memory'' represents the peak memory usage during inference with only the large VLM.   ``Large VLM with SGL peak memory'' indicates the peak memory usage during inference of the large VLM when using the proposed SGL method (guided by a 2B model). $\Delta$ is defined as the difference between ``Large VLM with SGL peak memory'' and ``Large VLM peak memory''. We report the ratio of $\Delta$ relative to ``Large VLM peak memory''.}
\label{app:tab_memory}
\end{table*}

\section{Visualization}
\begin{figure*}[!h]
    \centering
    \includegraphics[width=1\linewidth]{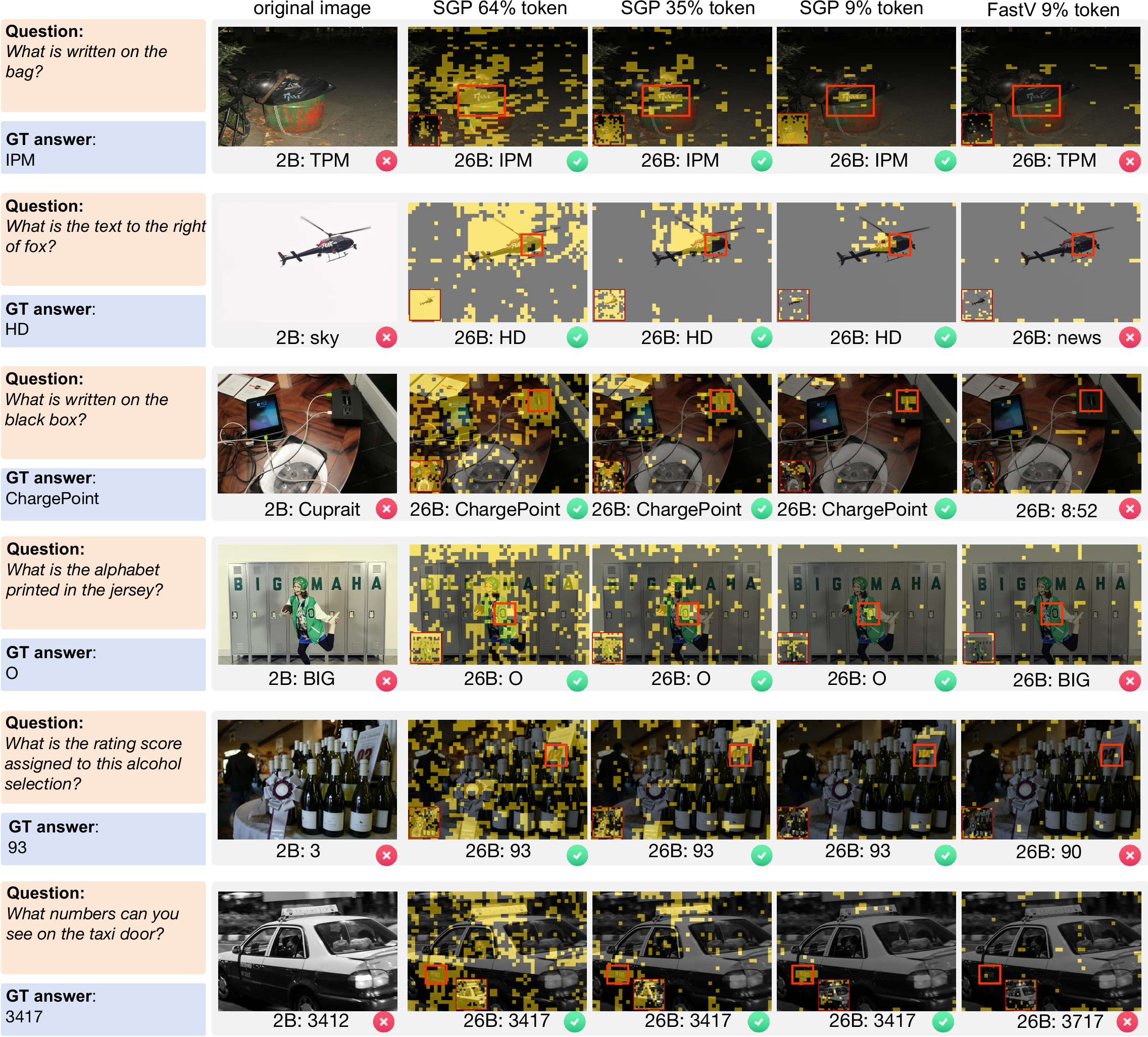}
    \vskip -0.1in
    \caption{\textbf{Additional visualization of SGP under different visual token retention ratios and answers.}  Visual tokens are pruned by 60\%, 80\%, and 95\% at the 19th, 9th, and 2nd layers of the large VLM of 26B, which comprises 48 layers. This results in average token retention ratios of 64\%, 35\%, and 9\%, respectively. Retained tokens are highlighted with \raisebox{0.25em}{\colorbox{important_token!50}{}}.  Thumbnails employed in InternVL are presented in the left corner. } 
    \label{app_fig:visualization}

\end{figure*}

In Figure~\ref{app_fig:visualization}, we provide additional visualizations of examples where the small VLM (2B) fails to produce correct predictions, while the large VLM (26B), with visual tokens pruned by SGP, successfully predicts the correct answers. Notably, in these cases, the large VLM with FastV~\cite{chen2024image} also fails.

\section{Model Descriptions}
The configurations of InternVL~\cite{chen2024far}, QWen2-VL~\cite{wang2024qwen2}, and LLaVa-OV~\cite{li2024llava} are comprehensively detailed in Tables~\ref{app:tab_intern}, \ref{app:tab_qwen2vl}, and \ref{app:tab_llava}, respectively.


\begin{table*}[!h]  
\renewcommand{\arraystretch}{1.0}
    \centering

\tablestyle{4.6pt}{1.2}

  \begin{tabular}{c  c  c c }

    \multirow{1}{*}{model name} & \multirow{1}{*}{language model}  & \multirow{1}{*}{vision encoder} & \multirow{1}{*}{checkpoint}\\
    \midrule[1.2pt]

InternVL2-1B & Qwen2‑0.5B~\cite{yang2024qwen2} & InternViT‑300M~\cite{chen2024internvl} & \href{https://huggingface.co/OpenGVLab/InternVL2-1B}{link} \\

InternVL2-2B & InternLM2‑chat‑1.8B~\cite{cai2024internlm2} & InternViT‑300M~\cite{chen2024internvl} & \href{https://huggingface.co/OpenGVLab/InternVL2-2B}{link} \\

InternVL2-4B & 	Phi‑3‑mini‑128k‑instruct~\cite{abdin2024phi} & InternViT‑300M~\cite{chen2024internvl} & \href{https://huggingface.co/OpenGVLab/InternVL2-4B}{link} \\

InternVL2-26B & InternLM2‑chat‑20B~\cite{cai2024internlm2} & InternViT‑6B~\cite{chen2024internvl} & \href{https://huggingface.co/OpenGVLab/InternVL2-26B}{link} \\

InternVL2-40B & Nous‑Hermes‑2‑Yi‑34B~\cite{Nous-Hermes-2-Yi-34B} & InternViT‑6B~\cite{chen2024internvl} & \href{https://huggingface.co/OpenGVLab/InternVL2-40B}{link} \\

InternVL2-76B & Hermes‑2‑Theta‑Llama‑3‑70B~\cite{Hermes-2-Theta-Llama-3-70B} & InternViT‑6B~\cite{chen2024internvl} & \href{https://huggingface.co/OpenGVLab/InternVL2-Llama3-76B}{link}

    \end{tabular}
    \caption{\textbf{Model descriptions of InternVL~\cite{chen2024far}} }
\label{app:tab_intern}
\end{table*}


\begin{table*}[!h]  
\renewcommand{\arraystretch}{1.0}
    \centering

\tablestyle{4.6pt}{1.2}

  \begin{tabular}{c  c  c c }

    \multirow{1}{*}{model name} & \multirow{1}{*}{language model}  & \multirow{1}{*}{vision encoder} & \multirow{1}{*}{checkpoint}\\
    \midrule[1.2pt]

Qwen2-VL-2B & Qwen2-1.5B~\cite{yang2024qwen2} & ViT~\cite{dosovitskiy2020image} &  \href{https://huggingface.co/Qwen/Qwen2-VL-2B-Instruct}{link}\\

Qwen2-VL-76B & Qwen2-72B~\cite{yang2024qwen2} &  ViT~\cite{dosovitskiy2020image} & \href{https://huggingface.co/Qwen/Qwen2-VL-72B-Instruct}{link} \\

    \end{tabular}
    \caption{\textbf{Model descriptions of QWen2-VL~\cite{wang2024qwen2}} }
\label{app:tab_qwen2vl}
\end{table*}


\begin{table}[!h]  
\renewcommand{\arraystretch}{1.0}
    \centering

\tablestyle{4.6pt}{1.2}

  \begin{tabular}{c  c  c c }

    \multirow{1}{*}{model name} & \multirow{1}{*}{language model}  & \multirow{1}{*}{vision encoder} & \multirow{1}{*}{checkpoint}\\
    \midrule[1.2pt]

LLaVa-OV-0.5B & Qwen2-0.5B~\cite{yang2024qwen2} & SigLIP~\cite{zhai2023sigmoid} &  \href{https://huggingface.co/lmms-lab/llava-onevision-qwen2-0.5b-ov}{link}\\

LLaVa-OV-72B & Qwen2-72B~\cite{yang2024qwen2} &  SigLIP~\cite{zhai2023sigmoid} & \href{https://huggingface.co/lmms-lab/llava-onevision-qwen2-72b-ov-sft}{link} \\

    \end{tabular}
    \caption{\textbf{Model descriptions of LLaVa-OV~\cite{li2024llava}}. }
\label{app:tab_llava}
\end{table}



\cleardoublepage
\twocolumn 



\end{document}